\documentclass[11pt]{article}

\usepackage[preprint]{acl}

\usepackage{times}
\usepackage{latexsym} 

\usepackage[T1]{fontenc}

\usepackage[utf8]{inputenc}

\usepackage{microtype}

\usepackage{inconsolata}

\usepackage{graphicx}
\usepackage{amssymb}
\usepackage{amsmath}
\usepackage{tikz}
\usepackage{booktabs}
\usepackage{tabularx}
\usepackage{subcaption}
\usepackage{tabularx}
\usepackage{booktabs}
\usepackage{float}
\usepackage{enumitem}
\usepackage{cuted}
\usepackage{caption}
\usetikzlibrary{arrows.meta, positioning, patterns, decorations.pathreplacing}
\title{Learning to Wait: Synchronizing Agents with the Physical World}

\author{
    \textbf{Yifei She\textsuperscript{1}, Ping Zhang\textsuperscript{1}, He Liu\textsuperscript{1}}, Yanmin Jia\textsuperscript{1}, Yang Jing\textsuperscript{1} \\
    \textbf{Zijun Liu\textsuperscript{3}, Peng Sun\textsuperscript{2}, Xiangbin Li\textsuperscript{1}, Xiaohe Hu\textsuperscript{1,}\thanks{\ \ Corresponding author.}} \\
    \textsuperscript{1}Infrawaves \\
    \textsuperscript{2}Shanghai Qiji Zhifeng Co., Ltd. \\
    \textsuperscript{3}Tsinghua University \\
    \texttt{\{sheyifei, zhangping, liuhe, jiayanmin, jingyang, lixiangbin, huxiaohe\}@infrawaves.com}\\
    \texttt{zj-liu24@mails.tsinghua.edu.cn, sunpeng@qijizhifeng.com}
}

\begin{document}
\maketitle
\begin{abstract}
Real-world agentic tasks, unlike synchronous Markov Decision Processes (MDPs), often involve non-blocking actions with variable latencies, creating a fundamental \textit{Temporal Gap} between action initiation and completion. Existing environment-side solutions, such as blocking wrappers or frequent polling, either limit scalability or dilute the agent's context window with redundant observations. In this work, we propose an \textbf{Agent-side Approach} that empowers Large Language Models (LLMs) to actively align their \textit{Cognitive Timeline} with the physical world. By extending the Code-as-Action paradigm to the temporal domain, agents utilize semantic priors and In-Context Learning (ICL) to predict precise waiting durations (\texttt{time.sleep(t)}), effectively synchronizing with asynchronous environment without exhaustive checking. Experiments in a simulated Kubernetes cluster demonstrate that agents can precisely calibrate their internal clocks to minimize both query overhead and execution latency, validating that temporal awareness is a learnable capability essential for autonomous evolution in open-ended environments.
\end{abstract}

\section{Introduction}
Constructing practical agentic training environments, such as Kubernetes (K8s) cluster sandboxes~\cite{ardebili2025kubeintellectmodularllmorchestratedagent}, presents a fundamental challenge: establishing precise temporal alignment between actions and their delayed feedback within a non-blocking asynchronous context. Unlike the synchronous Markov Decision Process (MDP) paradigm prevalent in frameworks like Gymnasium~\cite{towers2025gymnasiumstandardinterfacereinforcement}, real-world agent actions (e.g., \texttt{kubectl} commands) rarely trigger an immediate convergence of the environment state. Instead, they involve significant, variable temporal latency~\cite{bellinger2024dynamicobservationpoliciesobservation, pmlr-v139-biedenkapp21a}. This critical distinction involves a \textit{Temporal Gap} ($T_{\text{act}} \ll T_{\text{true}}$), which has been largely obscured by current research trends focusing on mathematical reasoning~\cite{deepseek-math-v2} or code generation~\cite{mai2025agentrlscalinglaw}. In such deterministic scenarios, interaction time is typically equivalent to task completion time ($T_{\text{act}} \approx T_{\text{true}}$), leaving the complexities of asynchronous temporal alignment effectively unexplored.

\begin{figure*}[h]
\centering
\includegraphics[width=\textwidth]{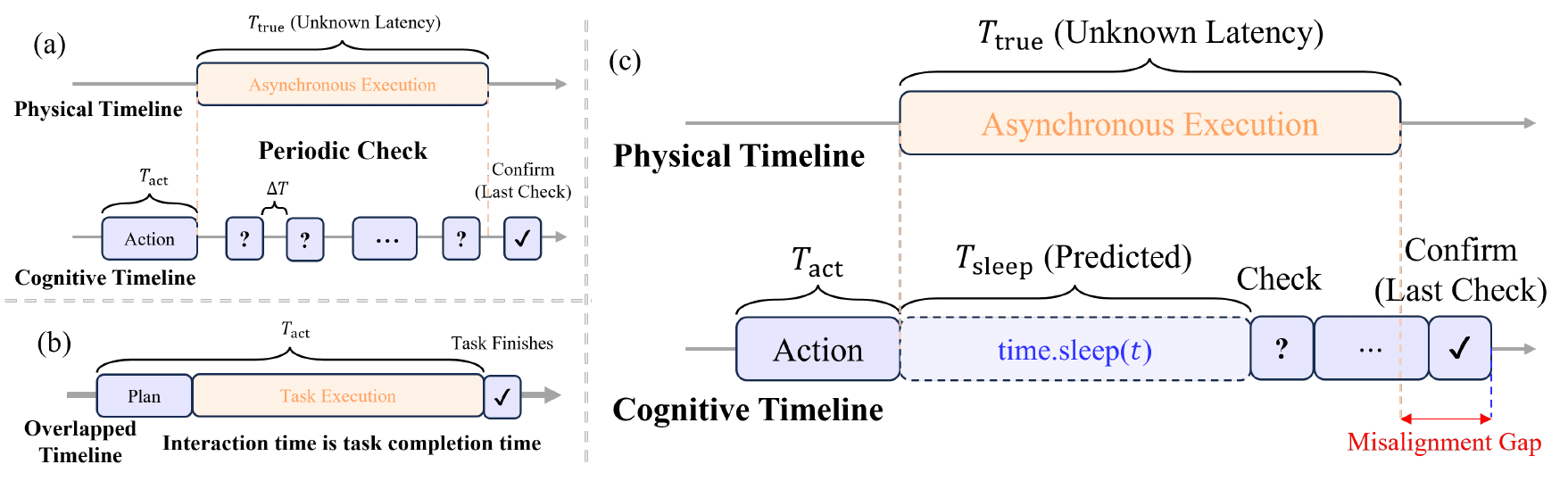}
\vspace{-2.0em}
\caption{\textbf{The Temporal Alignment Problem.} The asynchronous nature of real-world environments creates a discrepancy between the \textit{Physical Timeline} and the Agent's \textit{Cognitive Timeline}. While (a) \textbf{Periodic Check} forces alignment at prohibitively high query costs, (b) \textbf{General Agent Tasks} (e.g., Coding, Math) have obscured this challenge, as the agent's generation time naturally spans the task duration ($T_{\text{act}} \approx T_{\text{true}}$), leaving no temporal gap to manage, (c) \textbf{Our Approach} actively predicts an optimal $T_{\text{sleep}}$ to synchronize the agent's internal clock with physical latency, minimizing misalignment without redundant queries.}
\label{fig:timeline}
\vspace{-1.5em}
\end{figure*}

As illustrated in Figure~\ref{fig:timeline}, we abstract this challenge as an alignment problem between the agent's \textit{Cognitive Timeline} and the environment's \textit{Physical Timeline}. To bridge the discrepancy between event initiation and completion, two distinct paradigms exist: the \textbf{Environment-side Approach} and the \textbf{Agent-side Approach}. The Environment-side Approach relies on engineering wrappers to block execution or employs naive periodic polling. However, these methods are structurally flawed. Wrappers block the agent until tasks finish, forcing a hard coupling of the cognitive clock with the physical clock. Polling, while non-blocking, incurs non-negligible query costs~\cite{krueger2020activereinforcementlearningobserving} and, more critically, consumes the LLM's finite context window, which is a valuable resource. Filling the context window with redundant intermediate observations dilutes information density and degrades the signal-to-noise ratio, ultimately impairing the agent's reasoning capabilities~\cite{liu-etal-2024-lost}. The Agent-side Approach, conversely, acknowledges the independence of the physical world, requiring the agent to utilize its cognitive capabilities to predict a necessary waiting duration ($T_{\text{sleep}}$) to compensate for asynchronous latency.

Guided by the vision of Artificial General Intelligence (AGI) and the principles of the Scaling Law, we argue that the Agent-side perspective represents the more scalable and generalizable direction. Environment-side engineering shifts an unscalable maintenance burden onto developers by relying on hard-coded heuristics. Conversely, while mathematical estimators offer computational efficiency~\cite{nombebe2022fittinglomaxdistributioncomparison, Chen2018}, they remain fundamentally blind to the semantic nuances of latency. A statistical model indiscriminately averages latencies for identical actions (e.g., \texttt{docker pull}), failing to distinguish arguments; in contrast, an intelligent agent leverages semantic reasoning to distinguish how specific arguments differ in action duration~\cite{wei2023largerlanguagemodelsincontext}, recognizing the vast temporal disparity between pulling a lightweight base image and a massive CUDA image. Furthermore, by augmenting this intuition with external tools, such as querying search engines~\cite{li-etal-2025-search} to retrieve the specific size of a novel image, agents can transform blind waiting into active temporal planning, thereby achieving precise alignment with the physical world while minimizing query overhead.

To operationalize this alignment concept, we extend the \textbf{Code-as-Action} paradigm~\cite{pmlr-v235-wang24h} to the temporal domain. We empower the agent to actively align its cognitive clock by coding \texttt{time.sleep(t)}, thereby achieving synchronization with the physical timeline without blocking the execution flow. We validate this capability within an \textbf{Interleaved Action} framework, where the agent must execute a sequence of related yet distinct actions. This setting challenges the agent to move beyond learning a single global latency parameter, compelling it to simultaneously disentangle and calibrate the heterogeneous latency distributions of multiple actions. Leveraging In-Context Learning (ICL)~\cite{dong-etal-2024-survey}, the agent processes historical execution feedback, specifically the temporal alignment errors from previous episodes, to iteratively fine-tune its internal clock for each action. This approach demonstrates that an agent can transcend static memorization to achieve dynamic, multi-objective temporal calibration in non-stationary environments. More importantly, this ability to perceive, predict, and adapt to the environment's physical temporality is an indispensable prerequisite for building agents capable of continuous learning and self-evolution. Only by mastering the latency distribution of an asynchronous world can agents truly embark on their journey of autonomous evolution in open-ended environments.

\begin{figure*}[h]
    \centering
    \resizebox{0.86\textwidth}{!}{
        \begin{minipage}{\textwidth}
            \centering
            \begin{subfigure}[b]{0.48\textwidth}
                \centering
                \includegraphics[width=\textwidth]{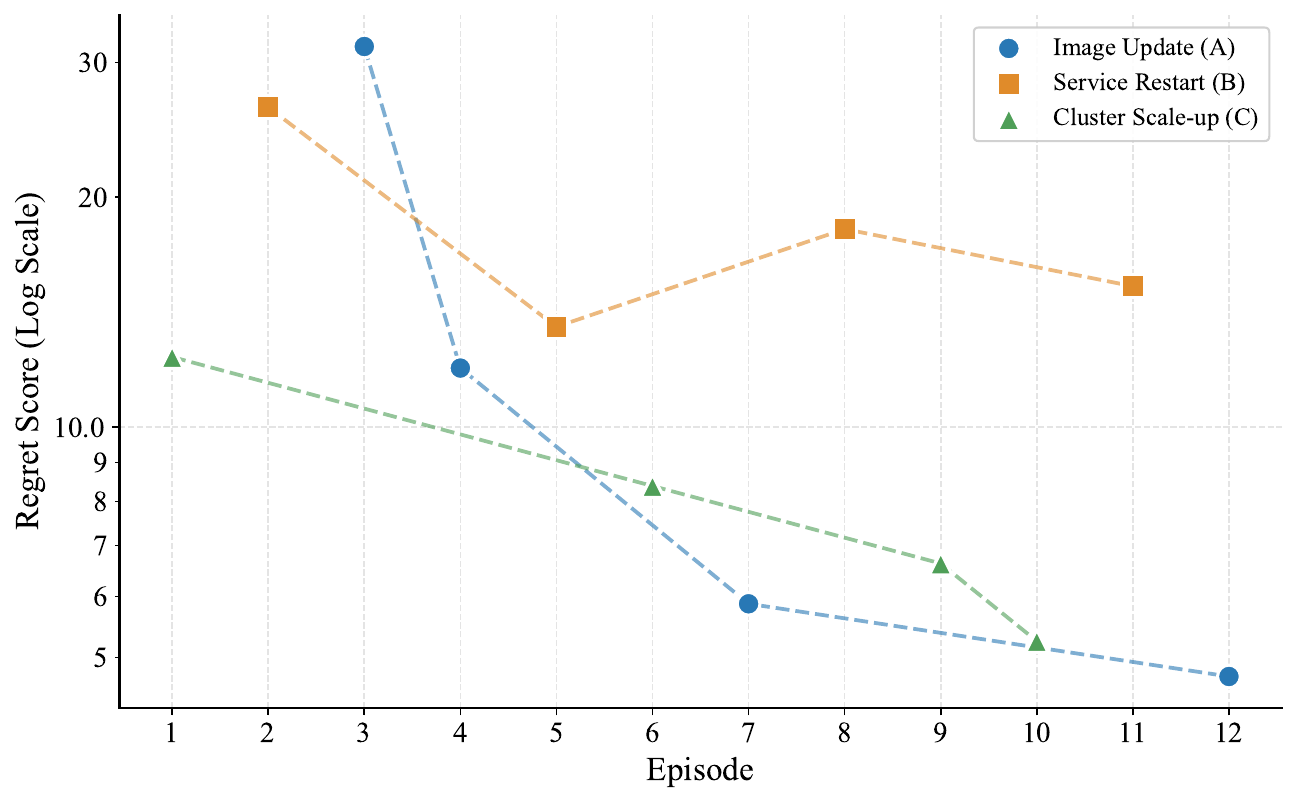}
                \caption{Gemini-3-Pro}
                \label{fig:regret_gemini}
            \end{subfigure}
            \hfill
            \begin{subfigure}[b]{0.48\textwidth}
                \centering
                \includegraphics[width=\textwidth]{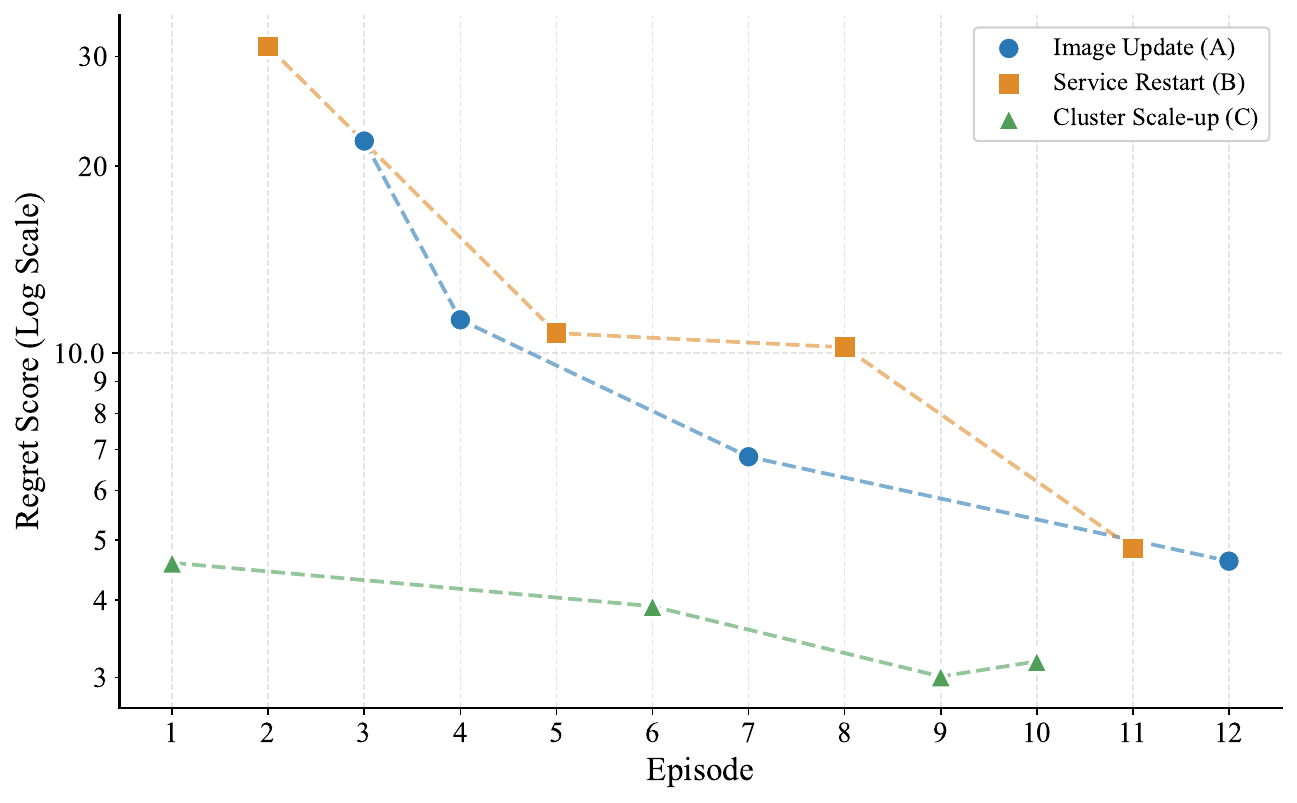}
                \caption{Claude-Sonnet-4.5}
                \label{fig:regret_claude}
            \end{subfigure}
            
            \vspace{0.3cm}

            \begin{subfigure}[b]{0.48\textwidth}
                \centering
                \includegraphics[width=\textwidth]{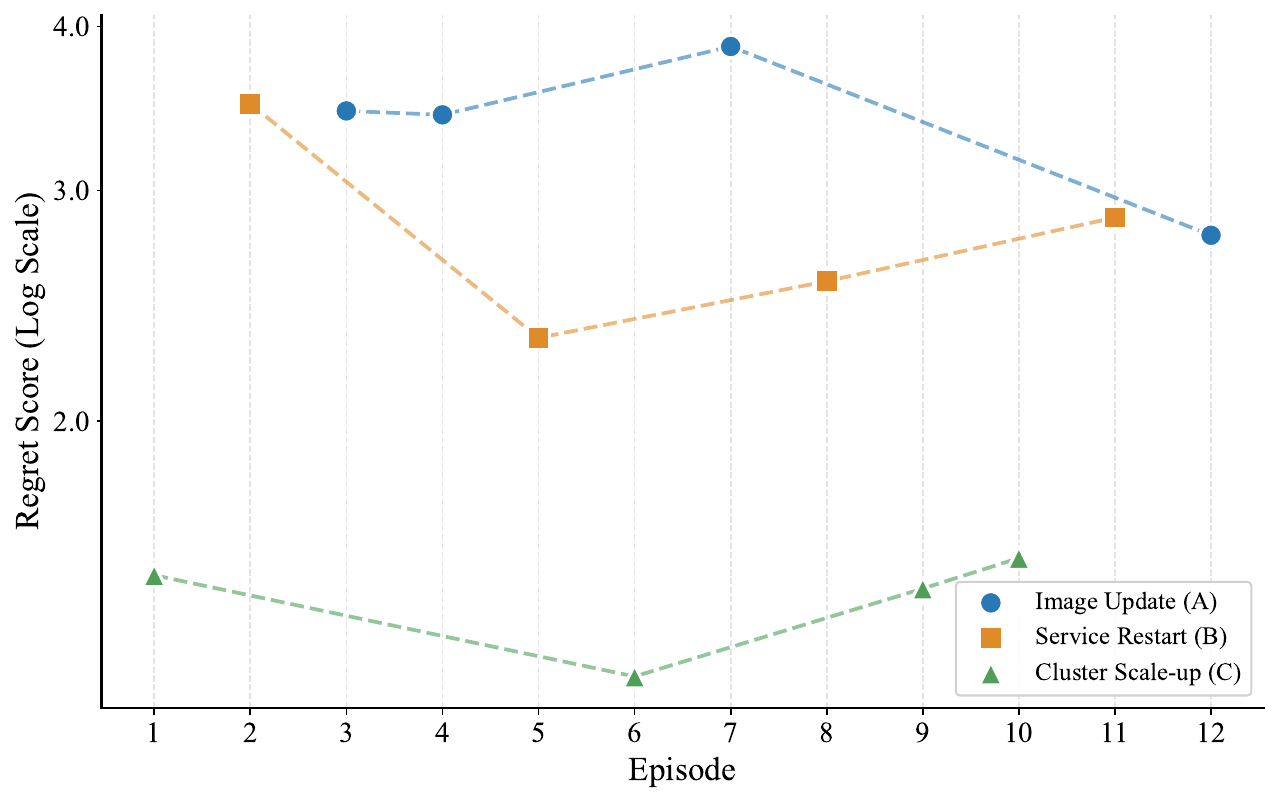}
                \caption{GPT-5.1}
                \label{fig:regret_gpt}
            \end{subfigure}
            \hfill
            \begin{subfigure}[b]{0.48\textwidth}
                \centering
                \includegraphics[width=\textwidth]{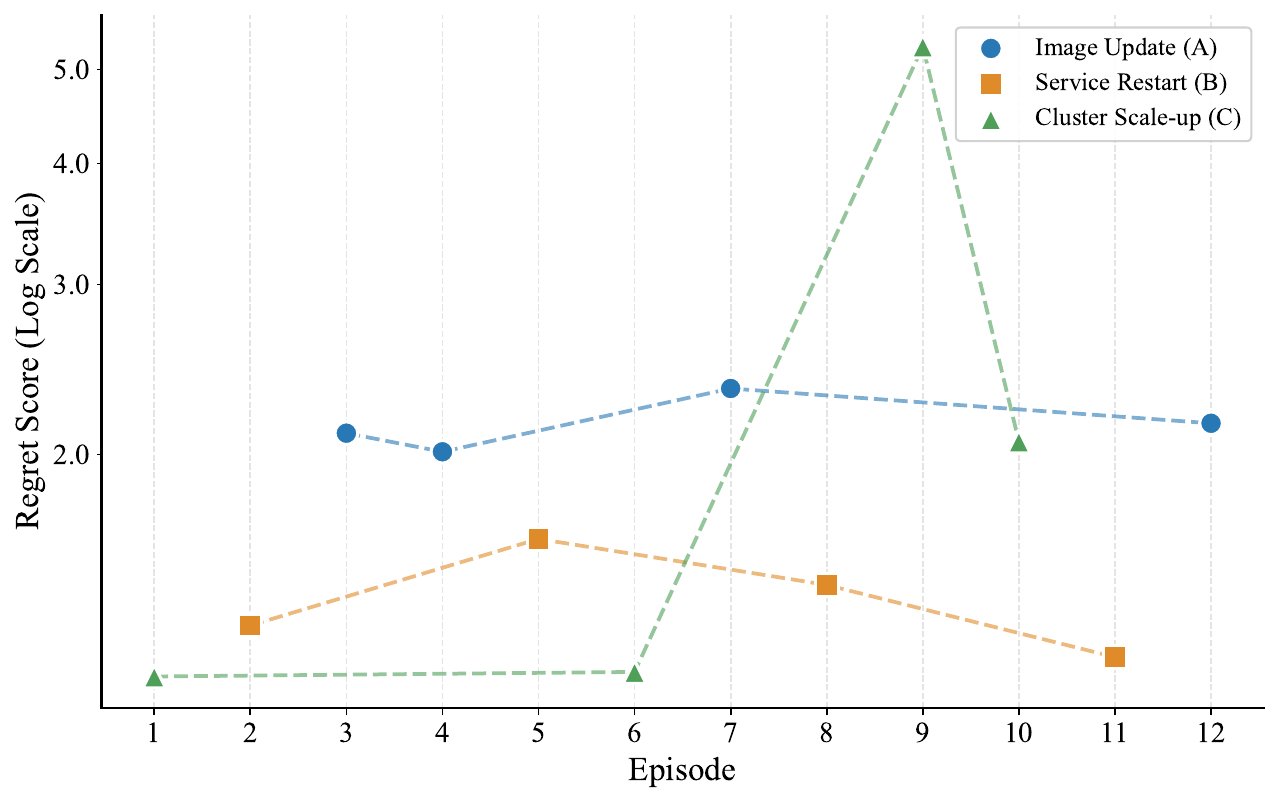}
                \caption{Qwen3-next-80B-A3B-Instruct}
                \label{fig:regret_qwen}
            \end{subfigure}
        \end{minipage}
    }
    \caption{A comparison of Regret Scores for four representative LLMs. The score quantifies the agent's efficiency, where a lower value indicates better performance.}
    \label{fig:regret_comparison}
    \vspace{-1.5em}
\end{figure*}

\section{Methodology}
We formulate the temporal adaptation challenge as a Repeated Temporal Game where the agent balances query costs against waiting latency. Given minimal state feedback, the agent must infer the distinct temporal characteristics of each action from execution history.

\subsection{Simulated Asynchronous Environment}
To simulate tasks with distinct temporal properties, we construct an environment where commands trigger background processes with stochastic latencies. Unlike atomic memoryless events, complex asynchronous tasks comprise multiple independent stages. Thus, we model the completion time $T_{\text{true}}$ using a \textbf{Gamma distribution}~\cite{Langaris1986} to capture this cumulative variance (details in Appendix~\ref{app:distribution visualization}). This formulation offers precise control over latency statistics for a realistic simulation.

The environment provides binary feedback $S \in \{\texttt{PENDING, DONE}\}$ without auxiliary progress hints. Operating on an already initiated task, the agent utilizes two tools: (1) Active Wait (\texttt{time.sleep(t)}) to calibrate its internal clock, and (2) Status Check (\texttt{env.check()}) to verify completion. The objective is to minimize check counts while aligning the wait duration with $T_{\text{true}}$.

\subsection{Inter-Episode History Feedback}
To leverage In-Context Learning (ICL), we structure the task as a multi-round repeated game. The prompt for the $k$-th episode includes an execution summary of the $(k-1)$-th episode. This summary contains the specific action performed, the number of checks and the total time consumed. Through this feedback loop, the agent incrementally refines its latency estimates for distinct actions, optimizing later predictions without updating parameters.

\begin{figure*}[ht]
    \centering
    \resizebox{0.90\textwidth}{!}{
        \begin{minipage}{\textwidth}
            \centering
            \begin{subfigure}[b]{0.48\textwidth}
                \centering
                \includegraphics[width=\textwidth]{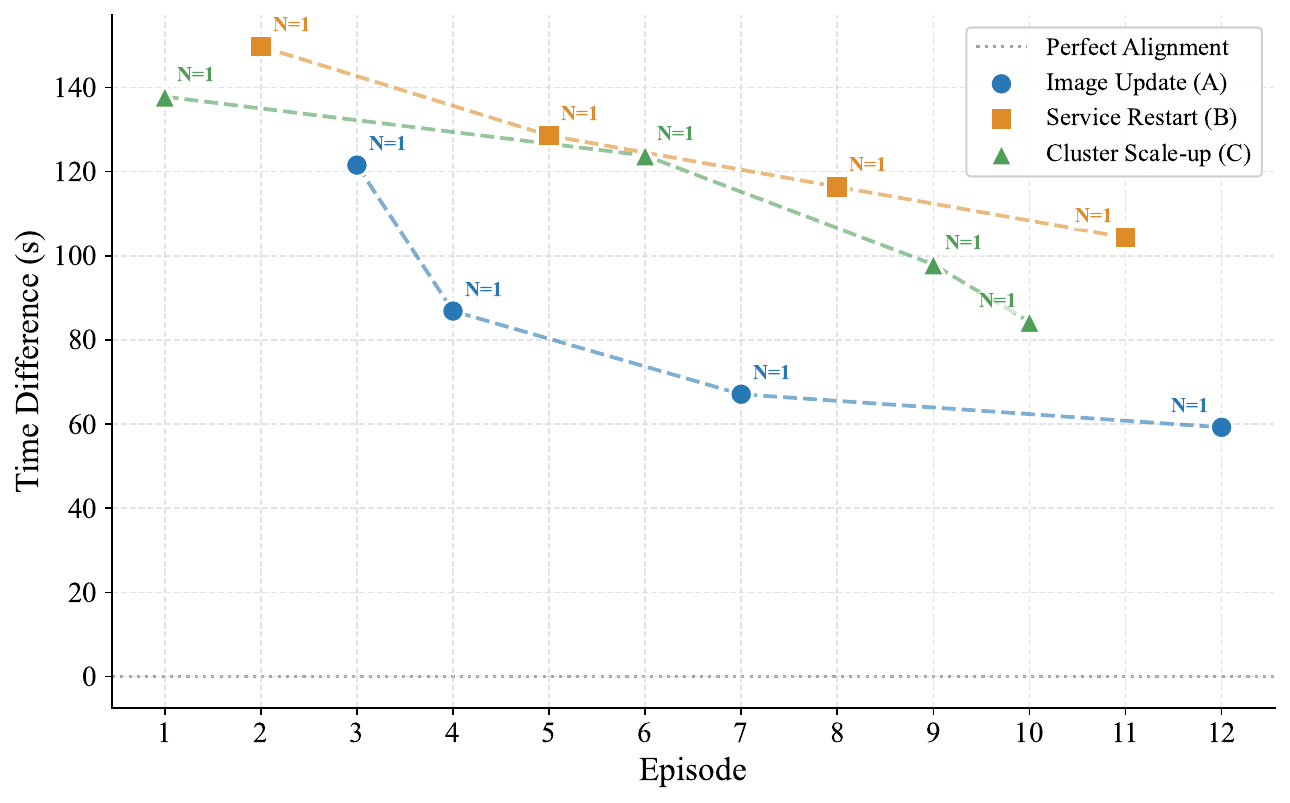}
                \caption{Gemini-3-Pro}
                \label{fig:diff_gemini}
            \end{subfigure}
            \hfill
            \begin{subfigure}[b]{0.48\textwidth}
                \centering
                \includegraphics[width=\textwidth]{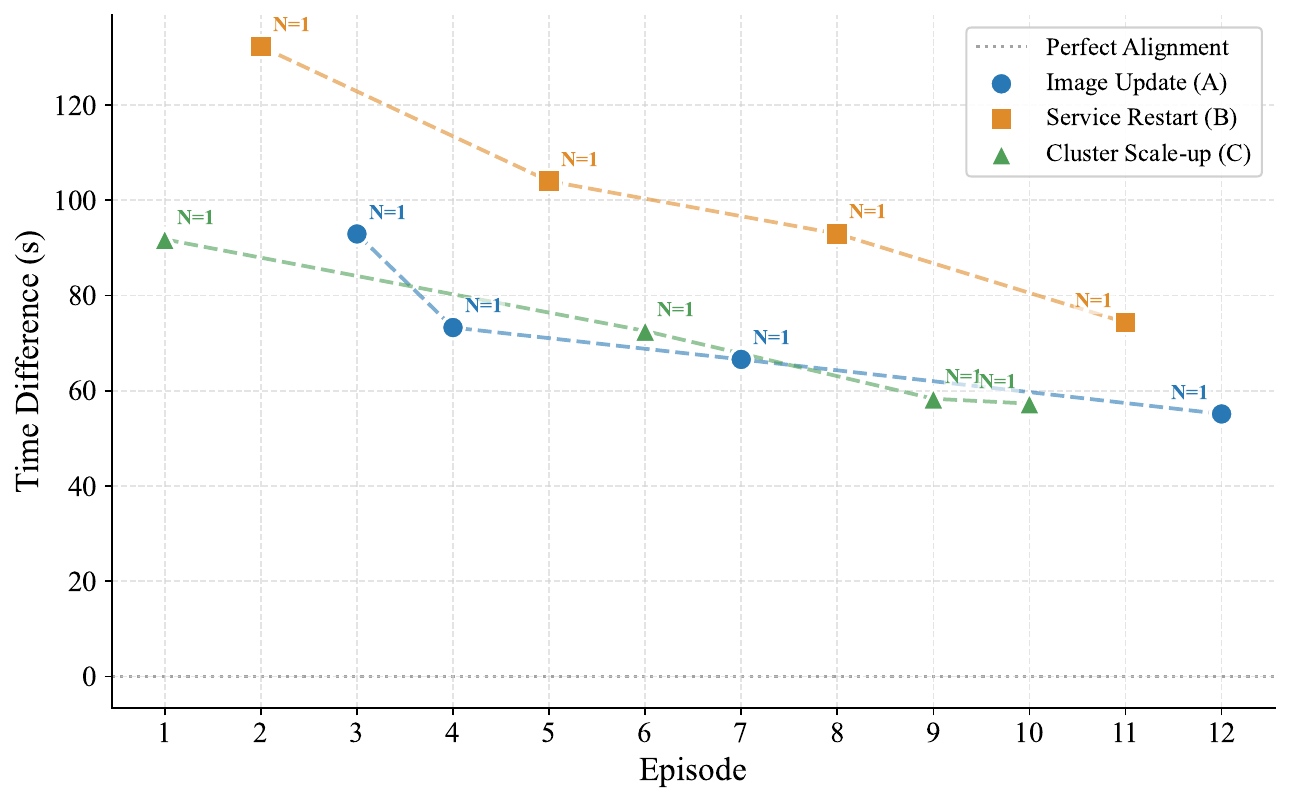}
                \caption{Claude-Sonnet-4.5}
                \label{fig:diff_claude}
            \end{subfigure}
            
            \vspace{0.3cm} 

            \begin{subfigure}[b]{0.48\textwidth}
                \centering
                \includegraphics[width=\textwidth]{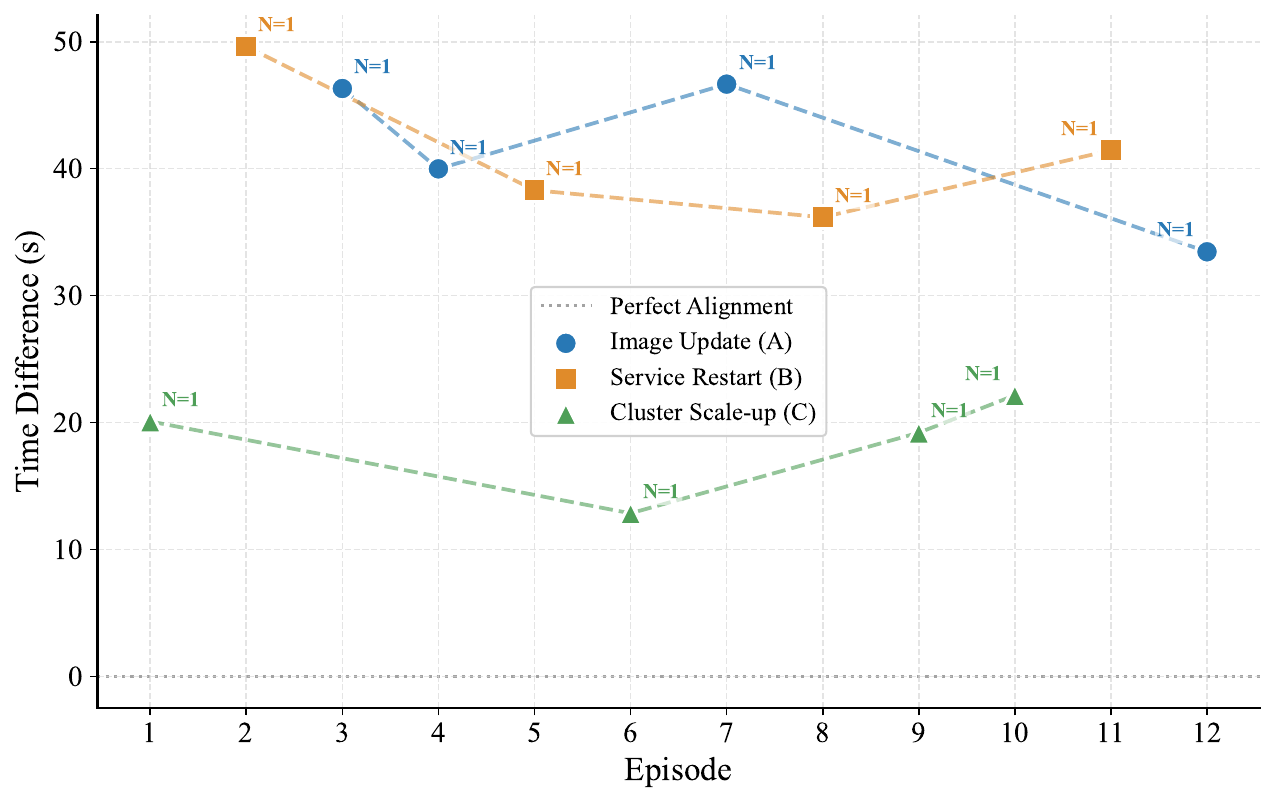}
                \caption{GPT-5.1}
                \label{fig:diff_gpt}
            \end{subfigure}
            \hfill
            \begin{subfigure}[b]{0.48\textwidth}
                \centering
                \includegraphics[width=\textwidth]{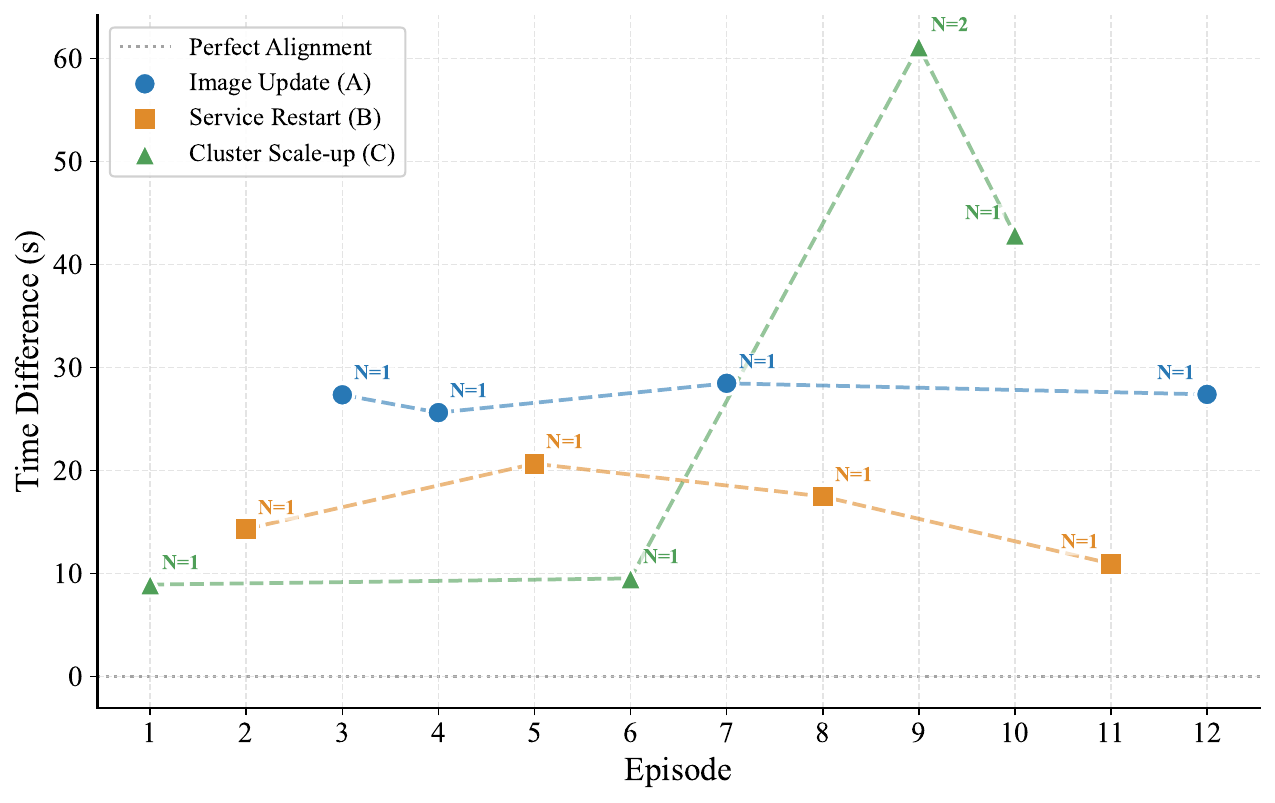}
                \caption{Qwen3-next-80B-A3B-Instruct}
                \label{fig:diff_qwen}
            \end{subfigure}
        \end{minipage}
    }
    \caption{Analysis of the temporal prediction error (\(T_{\text{confirm}} - T_{\text{true}}\)) over episodes. The annotation $N=1$ on the data points indicates the model only checks once.}
    \label{fig:diff_analysis}
    \vspace{-1.5em}
\end{figure*}

\section{Experiments and Analysis}
\label{sec:experiments and analysis}
Our experiments will test the hypothesis that LLMs leverage semantic priors for initial coarse-grained estimation and use historical feedback to calibrate actions, bypassing cold-start issues.

\subsection{Experimental Setup}
We simulate three distinct tasks modeled by independent Gamma distributions (shape $\alpha=20$, see Appendix~\ref{app:distribution visualization}) with varying means: a lightweight \textbf{Image Update} (Action A, $\mu=35$s), a medium \textbf{Service Restart} (Action B, $\mu=45$s), and a heavy \textbf{Cluster Scale-up} (Action C, $\mu=55$s). Full command details are in Appendix~\ref{app:command_details}. To simulate realistic constraints, completion times $T_{\text{true}}$ are bounded within intervals. The diversity requires the agent to distinguish between tasks using semantic cues, avoiding fitting a single global distribution.

To evaluate whether temporal adaptation is a universal capability across different models, we select four representative LLMs: \textbf{Gemini-3-Pro} (Reasoning-enhanced), \textbf{Claude-Sonnet-4.5} (Code-specialized), \textbf{GPT-5.1} (General SOTA), and \textbf{Qwen3-next-80B-A3B-Instruct} (Mid-sized open-weights). This selection allows us to analyze the impact of reasoning strength, coding proficiency, and model scale on temporal planning.

\subsection{Metric}
We define the \textbf{Regret Score} to quantify the deviation from optimal solution:
\begin{equation}
    \text{Score} = N_{\text{check}} \cdot \exp\left(\frac{T_{\text{confirm}} - T_{\text{true}}}{T_{\text{true}}}\right)
\end{equation}
where $N_{\text{check}}$ denotes the number of checks performed, and $(T_{\text{confirm}} - T_{\text{true}}) \geq 0$ represents the delay between the task actually becoming ready and the agent confirming it. A lower score indicates higher efficiency. This metric intentionally leverages the ground-truth $T_{\text{true}}$, because the focus of this work is to directly measure the agent's core capability to align its internal timeline with physical reality, not to provide a practical training recipe in online environments.\looseness=-1

\subsection{Results and Analysis}
Our experiments confirm that LLMs can indeed calibrate their \textit{Cognitive Timeline} to the physical world via In-Context Learning (ICL), though distinct behavioral patterns emerge across model architectures (Figure~\ref{fig:regret_comparison} and Figure~\ref{fig:diff_analysis}).

\paragraph{Dynamic Calibration via ICL.}
Reasoning-enhanced and code-specialized models, specifically \textbf{Gemini-3-Pro} and \textbf{Claude-Sonnet-4.5}, validate the core hypothesis. Both models exhibit a clear learning curve: they initiate with a conservative strategy (high safety buffer to ensure task completion), resulting in higher initial Regret Scores. However, leveraging historical feedback, they rapidly reduce the prediction error ($T_{\text{confirm}} - T_{\text{true}}$) over episodes. This demonstrates a genuine ability to treat time as an optimizable variable, dynamically trimming the wait buffer while maintaining the optimal checking frequency.\looseness=-1

\paragraph{Zero-shot Grounding vs. Rigid Heuristics.}
The behavior of \textbf{GPT-5.1} and \textbf{Qwen3-next} reveals a nuance between \textit{capability} and \textit{strategy}. 
\textbf{GPT-5.1} shows a flatter trend but maintains a consistently low Regret Score and Time Difference from the outset. This suggests superior zero-shot temporal grounding; its initial semantic priors were already precise enough to bypass the heavy calibration phase needed by others.
Conversely, while \textbf{Qwen3-next} achieves numerically competitive scores, a granular inspection of its trajectory reveals a failure in temporal adaptation. Instead of learning from context, the model defaults to a rigid heuristic (a static \texttt{time.sleep(60)}), which coincidentally covers the maximum latency ($\approx 55s$) and results in the clear plateau observed in Figure~\ref{fig:diff_analysis}. While effective in this specific bounded environment, this lack of plasticity implies a failure to generalize, highlighting the distinction between blind robustness and intelligent temporal alignment. \looseness=-1

\section{Conclusion}
This work bridges the Temporal Gap in asynchronous environments by empowering agents to actively manage their \textit{Cognitive Timeline}. We demonstrate that through the Code-as-Action paradigm, LLM can leverage semantic priors and historical feedback to align itself with physical latency dynamics. While model behaviors vary, the results confirm that temporal awareness is a learnable capability. This shift from environment-side engineering to agent-side adaptation not only minimizes query overhead but also establishes a fundamental prerequisite for self-evolving agents to efficiently interact with the non-blocking, variable-latency reality of real-world systems.

\newpage
\section*{Limitations}

While our results demonstrate the efficacy of Agent-side temporal alignment, we acknowledge two boundaries of the current study that invite future exploration. First, to ensure experimental control and reproducibility, we utilized a simulated asynchronous environment governed by Gamma distributions. While this statistically mirrors the multi-stage nature of real-world operations (e.g., container provisioning), it abstracts away low-level system noise (e.g., network jitter or preemption) that might introduce additional stochasticity in a live production cluster. Second, our evaluation focuses on the critical cold-start and rapid adaptation phases (first 12 episodes) of independent tasks. We prioritize this few-shot regime because real-world agents frequently encounter novel commands where long-term asymptotic convergence is less relevant than immediate efficiency. We acknowledge that the two-phase strategy in our prompt (predict a conservative delay, then gradually decrease) constitutes an expert-guided heuristic. While this validated ICL as an effective, light-weight calibration mechanism in a few-shot setting, a crucial future step is to explore how to train LLMs via RL to encode the latent temporal distribution into weights and further realize the self-evolving agent's adaptability in real cluster operations.


\bibliography{custom}

\appendix

\begin{table*}[h]
\centering
\caption{Details of the three simulated Kubernetes tasks. Each action is characterized by its semantic name, mean completion time ($\mu$), the full \texttt{kubectl} command, and the rationale for its expected latency, which is derived from the complexity of the underlying cluster operations.}
\label{tab:command_details}
\begin{tabularx}{\linewidth}{@{} c l c X @{}}
\toprule
\textbf{Action} & \textbf{Task Name} & \textbf{Mean Latency ($\mu$)} & \textbf{Command \& Rationale} \\
\midrule

A & Image Update & 35s &
    \small\texttt{kubectl set image deployment/webapp-frontend new-container=nginx:1.23.4}
    \vspace{1mm} \newline
    \textit{A lightweight operation that triggers a rolling update. The latency is dominated by control plane metadata changes, pod scheduling, and pulling a potentially cached image layer.} \\
\addlinespace 

B & Service Restart & 45s &
    \small\texttt{kubectl rollout restart statefulset/prometheus-db}
    \vspace{1mm} \newline
    \textit{A medium-weight operation that initiates a graceful, ordered restart of a stateful application. It is inherently slower as it must terminate and replace pods sequentially to maintain stability and data consistency.} \\
\addlinespace

C & Cluster Scale-up & 55s &
    \small\texttt{kubectl scale statefulset/etcd-cluster --replicas=5}
    \vspace{1mm} \newline
    \textit{A heavy-duty operation involving not just pod scheduling but also provisioning new persistent volumes (high I/O latency) and secure cluster quorum joining (network discovery and state sync).} \\

\bottomrule
\end{tabularx}
\end{table*}

\section{Distribution Visualization}
\label{app:distribution visualization}
\begin{figure}
    \centering
    \includegraphics[width=0.45\textwidth]{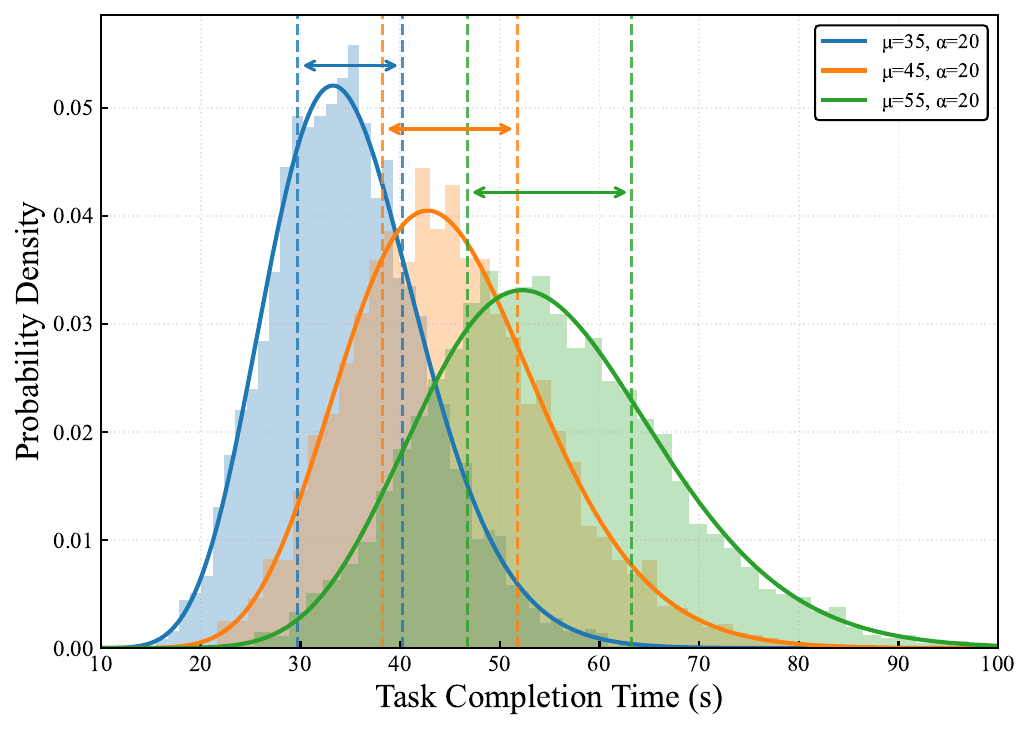}
    \caption{Probability density functions of the simulated task latencies. We model $T_{\text{true}}$ using Gamma distributions to capture the multi-stage nature of asynchronous operations.}
    \label{fig:gamma_distributions}
\end{figure}
Figure~\ref{fig:gamma_distributions} illustrates the probability density functions (PDF) of the task completion times ($T_{\text{true}}$) used in our simulated environment. We explicitly choose the \textbf{Gamma distribution} over the exponential distribution typically used in queuing theory. While a single atomic event (like a packet arrival) is memoryless and exponential, complex Kubernetes operations are composed of a sequence of dependent sub-stages (e.g., API validation $\to$ Pod termination $\to$ Image pulling $\to$ Container startup). The sum of multiple independent exponential variables follows a Gamma distribution.

\section{Kubernetes Task Command Details}
\label{app:command_details}

This section provides the full \texttt{kubectl} commands used in our experiments and a brief explanation of their expected operational latency in Table~\ref{tab:command_details}.

\section{Impact of Inference Latency on Temporal Calibration}
\label{app:inference_latency}

While a theoretical formulation might suggest explicitly subtracting the generation latency ($T_{\text{gen}}$) from the target wait time, we omit this step to maintain an end-to-end adaptive capability. Our experiments indicate that $T_{\text{gen}}$ exhibits high variance, rendering a static subtraction mechanism ineffective. Specifically, the agent might typically operate at two extremes: it either outputs the sleep command immediately, making $T_{\text{gen}}$ negligible, or engages in redundant reasoning where $T_{\text{gen}}$ alone suffices to cover the task duration, eliminating the need for additional waiting. Consequently, we treat the inference latency as an intrinsic component of the total system delay. This approach compels the agent to predict the residual wait time required after generation, thereby ensuring generalization across different models and inference infrastructures.

\section{Prompt Details}
\label{app:prompt_details}

Table \ref{tab:prompt_full} presents the exact prompt structure used in our experiments, formatted to illustrate the inputs and instructions provided to the agent.
\clearpage
\begin{table*}[h!]
    \centering
    \small
    \renewcommand{\arraystretch}{1.25}
    \begin{tabularx}{\linewidth}{lX}
        \toprule
        \textbf{Role} & \textbf{Content} \\
        \midrule
        
        \textbf{System} & You are a helpful AI agent. To solve the task, you must use the \texttt{execute\_python\_code} tool. Do not write code in your response directly. \\
        \midrule
        
        \textbf{User} & [CONTEXT] \par
        You are an intelligent agent operating in a Kubernetes environment. You have just executed the following kubectl command:
        \par
        \texttt{> > > \{kubectl command\}}
        \par
        Pay attention! You just edited the cluster and now need to wait for the command to complete. If you check too early, you'll find the task is still pending which will incur serious accidents. If you wait too long, you'll incur unnecessary delays. Your goal is to synchronize with the completion of this task effectively. The task is running in the background with a variable latency.
        
        \vspace{0.8em}
        [HISTORICAL TRAJECTORIES] \par
        To act, you MUST analyze the historical trajectories for this specific command and then use the \texttt{execute\_python\_code} tool. Provide the code to run as the 'code' parameter.
        \par
        \texttt{\{history\}}
        
        \vspace{0.8em}
        [AVAILABLE ACTIONS] \par
        \textbf{1. Check Status} \par
        - Description: Check if the target time has elapsed. This is the only way to successfully finish the task. \par
        - Tool Usage: Call \texttt{execute\_python\_code} with parameter \texttt{code="check()"}. \par
        - Output: "Status: PENDING..." (incomplete) or "Status: DONE..." (terminates episode).
        
        \vspace{0.5em}
        \textbf{2. Wait} \par
        - Description: Wait for a specific duration. \par
        - Tool Usage: Call \texttt{execute\_python\_code} with parameter \texttt{code="import time; time.sleep(seconds)"}. \par
        - Output: "Execution successful. Time has passed."
        
        \vspace{0.8em}
        [OPTIMIZATION GOAL] \par
        Your performance is evaluated by a \textbf{Regret Score} (Lower is better). The score is a combination of two factors:
        \begin{enumerate}[leftmargin=*, nosep]
            \item \textbf{Check Count (High Priority)}: Every check after the first one adds a large penalty. Your primary goal is to achieve \texttt{Check Count = 1}.
            \item \textbf{Wait Precision}: The time you wait should be as close as possible to the true (hidden) completion time. Waiting excessively long will also increase your Regret Score.
        \end{enumerate}
        
        \vspace{0.8em}
        [TWO-PHASE STRATEGY] \par
        Follow this procedure to make your decision:
        \par
        \textbf{Phase 1: Establish a Safe Baseline} \par
        - If this is the first time you see this specific command, you have no history. Make a conservative first guess based on your semantic prior knowledge. A long wait is better than failing. Goal: get the first data point with \texttt{Check Count = 1}.
        
        \vspace{0.3em}
        \textbf{Phase 2: Cautious Optimization} \par
        - Once you consistently achieve \texttt{Check Count = 1}, analyze the history for the \textit{current command type only}. \par
        - Find your last successful wait time (\texttt{Last\_Wait}). \par
        - Propose a new wait time that is \textbf{slightly less than \texttt{Last\_Wait}} (e.g., 10-20\% reduction) but not so low that you risk failing.
        
        \vspace{0.8em}
        [SYSTEM INSTRUCTIONS] \par
        At each step, you must decide whether to \texttt{wait} or \texttt{check} and call the \texttt{execute\_python\_code} tool accordingly. Use \texttt{wait} to approach the target time, and \texttt{check} only when you are confident the time has elapsed. \\

        \midrule
        \textbf{History Format} & 
        \textbf{Episode $i$:} \par
        \textbf{Command} = \texttt{'kubectl command'}, \newline
        \textbf{Your Executed Sleep Time} = $t$s, \newline
        \textbf{Check Count} = $N$ \\
        \bottomrule
    \end{tabularx}
    \caption{The complete prompt template used for the Agent-side Approach. The prompt provides the agent with the semantic context, historical execution data, and a clear two-phase strategy to minimize the Regret Score.}
    \label{tab:prompt_full}
\end{table*}
\clearpage

\begin{strip}
    \centering
    \captionsetup{type=figure} 
    
    \begin{subfigure}[b]{0.48\textwidth}
        \centering
        \includegraphics[width=\linewidth]{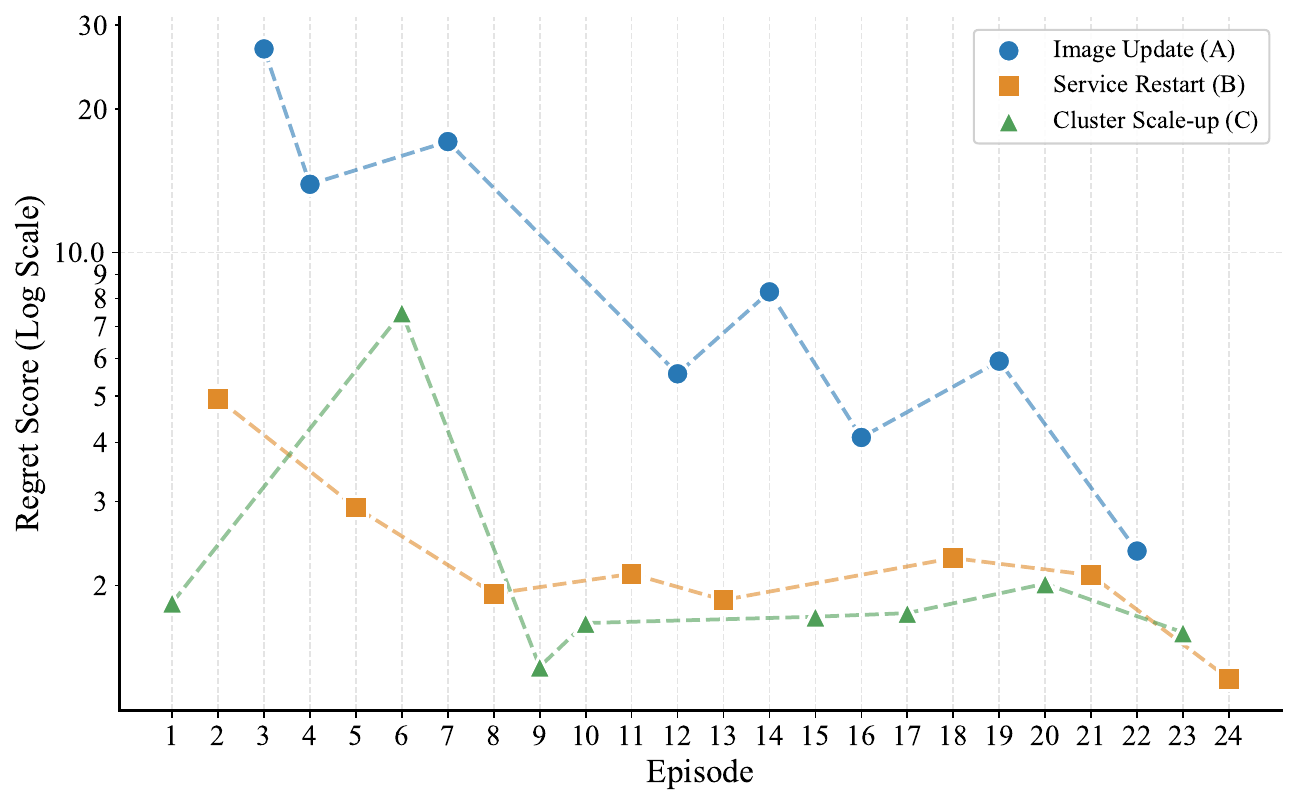}
        \caption{Regret Score Trajectory}
        \label{fig:kimi_regret}
    \end{subfigure}
    \hfill
    \begin{subfigure}[b]{0.48\textwidth}
        \centering
        \includegraphics[width=\linewidth]{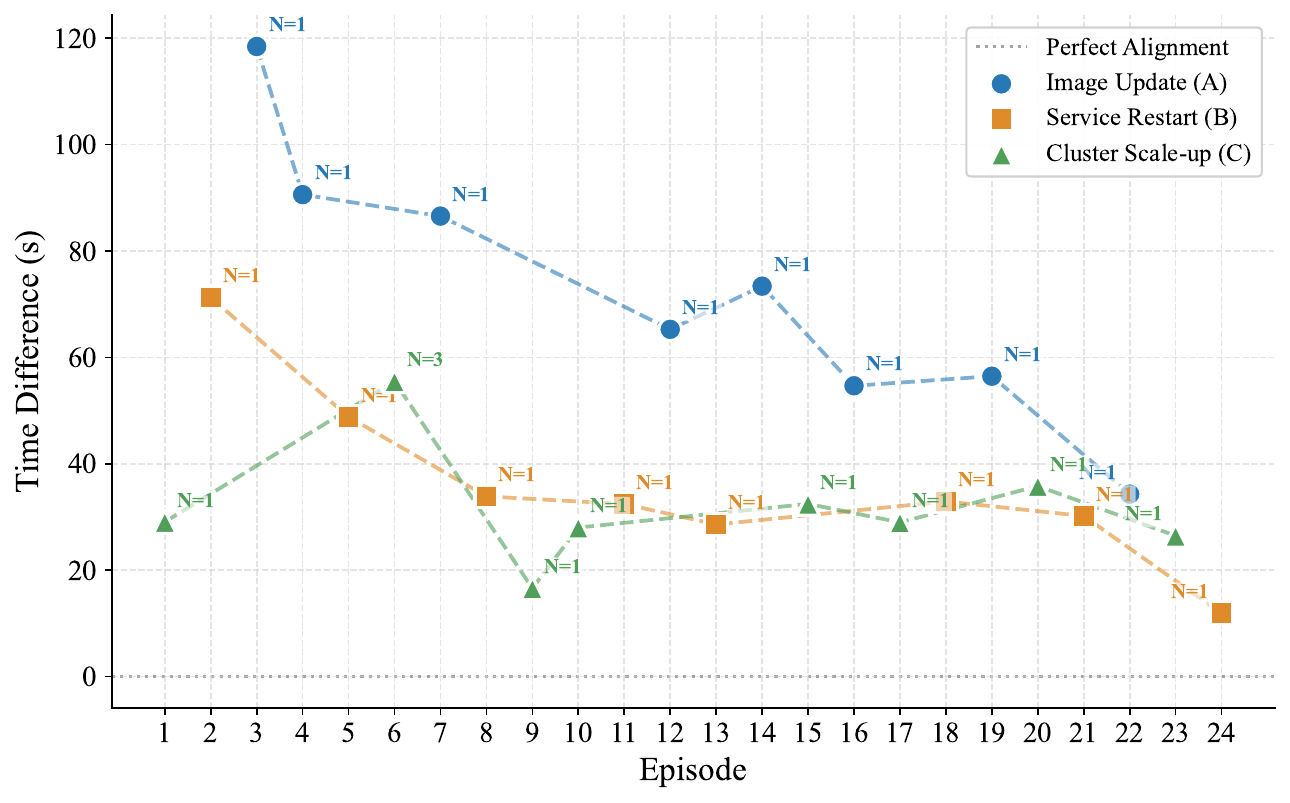}
        \caption{Time Difference Trajectory}
        \label{fig:kimi_time_diff}
    \end{subfigure}
    
    \captionof{figure}{\textbf{Temporal Adaptation Dynamics of kimi-k2-0905.} 
    (a) The significant reduction in Regret Score demonstrates the agent's rapid convergence towards an optimal checking strategy. 
    (b) The Time Difference metric reveals how the agent minimizes the \textit{Temporal Gap}, effectively synchronizing its \textit{Cognitive Timeline} with the asynchronous physical latency. Annotations ($N=x$) indicate the check count for specific episodes.}
    \label{fig:kimi_experiment_results}
\end{strip}

\section{Evolutionary Trajectory of Alignment}
\label{app:kimi_analysis}

To provide a granular perspective on how Large Language Models align their internal \textit{Cognitive Timeline} with the external \textit{Physical Timeline}, we conducted an extended long-horizon evaluation (24 episodes) utilizing \textbf{kimi-k2-0905}. This qualitative case study empirically validates the \textit{Two-Phase Strategy} proposed in our methodology. By analyzing the Chain-of-Thought (CoT) reasoning traces, we demonstrate how the agent transitions from utilizing generalized semantic priors to performing precise, history-driven calibration.

\paragraph{Phase 1: Zero-Shot Temporal Grounding}
In the cold-start phase, the agent lacks historical execution data. Consequently, it must rely on \textit{Zero-Shot Temporal Grounding}—inferring physical latency directly from the semantic complexity of the command. As shown in Table~\ref{tab:cot_reasoning}, when the agent first encounters the \textit{Cluster Scale-up} task (Episode 1), it identifies key terms such as "etcd" and "distributed consensus." Recognizing that maintaining state consistency is time-consuming, it sets a conservative baseline of 60 seconds to ensure task success. Importantly, this reasoning is context-aware rather than static. In Episode 2, when facing a \textit{Service Restart}, the agent explicitly compares it to the previous scaling task. It reasons that a "restart" involves a two-step process (termination followed by creation) and is therefore computationally heavier than simple scaling. Based on this semantic derivation, it proactively increases the safety buffer to 90 seconds. This demonstrates that the agent can construct a hierarchical mental model of task latency solely through semantic reasoning.

\paragraph{Phase 2: Adaptive Calibration via Feedback}
Once historical data becomes available, the agent shifts its strategy to \textit{Adaptive Calibration}. The objective changes from \textbf{ensuring success} (maximizing safety) to \textbf{minimizing regret} (optimizing efficiency). By analyzing the \textit{Inter-Episode History Feedback}, the model detects trends in the execution time. For example, in the later stages of the experiment (Episode 24), the agent observes a sequence of successful executions where the wait time decreased from 90s to 45s. Recognizing that the system has stabilized, the agent abandons the initial conservative estimate and applies a decay factor (approximately 10\%) to the wait time. This behavior confirms that the agent can treat time as an optimizable variable. Through In-Context Learning, the agent effectively fine-tunes its internal clock to match the stochastic reality of the environment, achieving asymptotic convergence without updating its parameters.

\begin{table*}[p]
\centering
\small
\renewcommand{\arraystretch}{1.25}
\setlength{\tabcolsep}{8pt}
\begin{tabular}{p{0.04\linewidth} p{0.14\linewidth} p{0.68\linewidth} p{0.06\linewidth}}
\toprule
\textbf{Ep} & \textbf{Task Type} & \textbf{Chain-of-Thought Reasoning (Compressed by LLM)} & \textbf{$T_{sleep}$} \\ 
\midrule
\multicolumn{4}{c}{\textit{\textbf{Phase 1: Zero-Shot Temporal Grounding (Establishing Safe Baselines)}}} \\
\midrule

1 & Cluster Scale-up & 
I need to analyze the situation and make a decision about waiting for the command to complete. Since I have no historical trajectories for this specific command, I need to establish a safe baseline. The command is scaling a StatefulSet, which involves bringing up new pods in sequence with proper ordering and stability checks. \textbf{For an `etcd` cluster, this is particularly important as `etcd` requires careful pod management due to its distributed consensus algorithm.} While a safe conservative estimate would be around 90-120 seconds, I will start with a moderately conservative approach. I will wait 60 seconds as my initial safe baseline to give the StatefulSet controller enough time to create and stabilize the new pods. 
& 60s \\ \hline

2 & Service Restart & 
This is the first time I am encountering the `kubectl rollout restart` command in the historical trajectory. Looking at the reference point: Episode 1 (`scale`) took 60s. However, \textbf{a rollout restart operation is more complex}—it needs to gracefully terminate existing pods and create new ones. This process depends on factors like the number of replicas, pod termination grace periods, image pull times, and resource availability. Since this is my first attempt and I need to establish a safe baseline, I will choose a conservative wait time that is longer than the 60s used in the reference scaling example. I will wait 90 seconds as a reasonable, conservative first guess. 
& 90s \\ \hline

3 & Image Update & 
The command is `kubectl set image`. Looking at the history, I see Episode 1 (Scale, 60s) and Episode 2 (Restart, 90s). I do not have history for this exact command type, but general Kubernetes knowledge suggests that updating a deployment requires rolling out new pods and terminating old ones. \textbf{This process is structurally more similar to a `rollout restart` than a simple `scale` operation.} The process usually takes longer than simple scaling operations depending on the deployment size. Given no specific history, I need a safe baseline. I will use a conservative estimate significantly longer than the historical times (60s and 90s) to ensure I achieve `Check Count = 1`. I will wait 120 seconds. 
& 120s \\

\midrule
\multicolumn{4}{c}{\textit{\textbf{Phase 2: Adaptive Calibration (Optimization \& Convergence)}}} \\
\midrule

4 & Image Update & 
Looking at the historical trajectories, I can see that this exact command was executed in Episode 3 with a wait time of 120.0s and completed successfully (Check Count = 1). This means I have perfect history for this specific command. Following Phase 2 (Cautious Optimization), I should try to reduce the wait time from the previous 120.0s by a safe margin to improve efficiency. \textbf{A 10\% reduction seems reasonable—that is $120s \times 0.9 = 108s$.} This represents a small but meaningful improvement in precision while maintaining a high confidence of success. I will wait for 108 seconds. 
& 108s \\ \hline

6 & Cluster Scale-up & 
Looking at the historical trajectories, I can see this exact command `kubectl scale` was executed in Episode 1 with a wait time of 60.0s and successful completion (Check Count = 1). Since I have history for this specific command and it was successful, \textbf{I should use this as my baseline.} I do not need to reduce the time yet because I only have one data point and I want to ensure success. My priority is to guarantee the task completes rather than optimization at this early stage. I will wait the same 60s to guarantee the task completes. 
& 60s \\ \hline

11 & Service Restart & 
Looking at the historical data for `rollout restart`, I can see a clear trend: Episode 2 (90s), Episode 5 (80s), and Episode 8 (70s). All were successful with Check Count = 1. \textbf{The trend shows the wait times decreasing from 90s $\to$ 80s $\to$ 70s.} The most recent successful wait was 70s. Since I am consistently achieving success but want to optimize precision, I will reduce the wait time by a small margin (about 15\%) from the last successful attempt. Calculation: $70s \times 0.85 \approx 60s$. 
& 60s \\ \hline

20 & Cluster Scale-up & 
This is a repeated operation. From the history, I see the following wait times: Episodes 1, 9, 10, 15, 17 used various times ranging from 58-65s, all with Check Count = 1. Episode 6 (75s) had Check Count = 3, which indicates the wait was too short (likely an outlier). \textbf{The most recent successful executions (Episodes 15 and 17) used 65s and 58s respectively.} Since I want to cautiously optimize while maintaining Check Count = 1, I will use a wait time that is slightly less than the most recent successful time. I will wait 60 seconds, which is a balanced choice based on the historical data. 
& 60s \\ \hline

24 & Service Restart & 
Looking at the historical trajectories, I see the following execution times: 90s (Ep 2) $\to$ 80s $\to$ 70s $\to$ 60s $\to$ 55s $\to$ 50s $\to$ 45s (Ep 21). \textbf{I notice a clear trend of decreasing wait times over the episodes, with the most recent successful execution taking only 45s.} This suggests the system has been optimized and is now working more efficiently. Following the optimization strategy, I should reduce the wait time slightly from the last successful execution. I will decrease by about 10\% from 45s to approximately 40s. 
& 40s \\ 

\bottomrule
\end{tabular}
\caption{The partial CoT trajectories of kimi-k2-0905.}
\label{tab:cot_reasoning}
\end{table*}

\end{document}